\documentclass[journal,compsoc]{IEEEtran}

\usepackage{graphicx}
\usepackage{algorithm}
\usepackage{algorithmic}
\usepackage{array}

\usepackage{longtable,array,ragged2e}
\newcolumntype{P}[1]{>{\RaggedRight\arraybackslash\hspace{0pt}}p{#1}}
\usepackage{amsthm}

\usepackage{url}
\usepackage{amsmath}
\usepackage[cmintegrals]{newtxmath}
\usepackage[table]{xcolor}
\usepackage{supertabular}
\usepackage{makecell}
\usepackage{multicol}

\usepackage[]{amssymb}
\usepackage[dvips]{epsfig}
\usepackage{lscape}
\usepackage{multirow}
\usepackage{color}
\usepackage{diagbox}
\usepackage{array}%?????
\usepackage[normalem]{ulem}
\usepackage{subcaption}
\usepackage{ltxtable}
\usepackage{tabularx}
\usepackage{rotating}
\usepackage{verbatim}

\newcommand\so{\bgroup\markoverwith{\textcolor{red}{\rule[0.5ex]{2pt}{0.4pt}}}\ULon}

\usepackage{xcolor}
\usepackage{colortbl}

\begin{document}

\title{The Frontier of Data Erasure: \\ Machine Unlearning for Large Language Models}


\author{Youyang Qu,~\IEEEmembership{Member,~IEEE,}
        Ming Ding,~\IEEEmembership{Senior Member,~IEEE,}
        Nan Sun,~\IEEEmembership{Member,~IEEE,} \\
        Kanchana Thilakarathna@sydney.edu.au,~\IEEEmembership{Senior Member,~IEEE,} \\
        Tianqing Zhu,~\IEEEmembership{Senior Member,~IEEE,} 
        and Dusit Niyato,~\IEEEmembership{Fellow,~IEEE}%<-this % stops a space
\IEEEcompsocitemizethanks{\IEEEcompsocthanksitem Youyang Qu and Ming Ding are with Data61, Commonwealth Scientific and Industrial Research Organization, Australia. Email: \{youyang.qu, ming.ding\}@data61.csiro.au.

Nan Sun is with the School of Systems \& Computing, University of New South Wales, Australia. Email: nan.sun@adfa.edu.au. 

Kanchana Thilakara is with the School of Computer Science, University of Sydney, Australia. Email: kanchana.thilakarathna@sydney.edu.au. 

Tianqing Zhu is with the Faculty of Data Science, City University of Macau. Email: tianqing.zhu@uts.edu.au. 

Dusit Niyato is with the College of Computing \& Data Science, Nanyang Technological University. Email: dniyato@ntu.edu.sg. 


}\protect\\% <-this % stops an unwanted space
\iffalse \thanks{Manuscript received April 19, 2005; revised August 26, 2015.} \fi
}

\markboth{IEEE Computer Magazine}%
{Shell \MakeLowercase{\textit{et al.}}: Bare Demo of IEEEtran.cls for IEEE Transactions on Magnetics Journals}



% for Transactions on Magnetics papers, we must declare the abstract and
% index terms PRIOR to the title within the \IEEEtitleabstractindextext
% IEEEtran command as these need to go into the title area created by

\IEEEtitleabstractindextext{%
\begin{abstract}
Large Language Models (LLMs) are foundational to AI advancements, facilitating applications like predictive text generation. Nonetheless, they pose risks by potentially memorizing and disseminating sensitive, biased, or copyrighted information from their vast datasets. Machine unlearning emerges as a cutting-edge solution to mitigate these concerns, offering techniques for LLMs to selectively discard certain data. This paper reviews the latest in machine unlearning for LLMs, introducing methods for the targeted forgetting of information to address privacy, ethical, and legal challenges without necessitating full model retraining. It divides existing research into unlearning from unstructured/textual data and structured/classification data, showcasing the effectiveness of these approaches in removing specific data while maintaining model efficacy. Highlighting the practicality of machine unlearning, this analysis also points out the hurdles in preserving model integrity, avoiding excessive or insufficient data removal, and ensuring consistent outputs, underlining the role of machine unlearning in advancing responsible, ethical AI.
\end{abstract}

}

\maketitle

\IEEEdisplaynontitleabstractindextext

% \fi
\IEEEpeerreviewmaketitle

\section{Introduction}

\IEEEPARstart{L}{arge} Language Models (LLMs) have transformed the landscape of AI, 
offering remarkable abilities in enhancing and generating human language text. 
However, 
their strength, 
derived from vast datasets, 
can become a liability due to privacy concerns, accuracy limitations, copyright infringement issues, 
and the potential propagation of societal biases~\cite{singhal2023large}.
A notable instance is the lawsuit filed by the New York Times against OpenAI and Microsoft for using its copyrighted content in training their GPT models, 
igniting a controversial debate on the application of fair use rules to LLM training and spotlighting an urgent need for data erasure mechanisms in LLMs.\footnote{https://www.theverge.com/2023/12/27/24016212/new-york-times-openai-microsoft-lawsuit-copyright-infringement} 
%This has ushered in a critical need for effective unlearning mechanisms within these models. 

In this light, 
a concept of machine ``unlearning'' has been recently proposed to realize data erasure~\cite{cao2015towards}. 
Machine unlearning is applied to LLMs due to a few challenges, 
including the need for fast retraining or fine-tuning, 
removing the impact of outdated, copyrighted, and false data, etc. 
A recent paper discussed the ``Right to be forgotten'' in LLMs~\cite{zhang2023right}, 
where the authors provided some implications of laws like European General Data Protection Regulation (GDPR)\footnote{https://gdpr-info.eu/}, and classified the methods into exact and approximate machine unlearning in a general setting. However, it lacks a dedicated taxonomy, challenges analysis, evaluation, as well as insights derived from the latest research.  

In this paper, 
we categorize machine unlearning for LLMs into two streams. 
The first stream primarily focuses on unlearning unstructured data like certain knowledge within language models. 
This encompasses removing or modifying specific information, narratives, or sentences the model has been trained on~\cite{jang2022knowledge}. 
The objective is to make the LLMs ``forget'' certain learned content, 
which might be sensitive, copyrighted, or incorrect, 
without compromising their general linguistic capabilities~\cite{eldan2023s}. 
Such an approach addresses legal and ethical concerns, 
particularly in scenarios where LLMs inadvertently reproduce copyrighted material or retain sensitive personal data~\cite{karamolegkou2023copyright}.

The second research stream targets unlearning structured data to enhance the classification abilities of LLMs. 
In this context, 
unlearning is adopted to refine the model’s decision-making processes, 
reducing biases and improving its interpretative accuracy~\cite{pawelczyk2023context}. 
This research branch is particularly significant in creating fairer, 
more balanced models that do not perpetuate existing societal prejudices or skewed perspectives.

\begin{figure*}[!htbp]
\centering
\subfloat[Unlearning structured data]{
\includegraphics[width=3.2in]{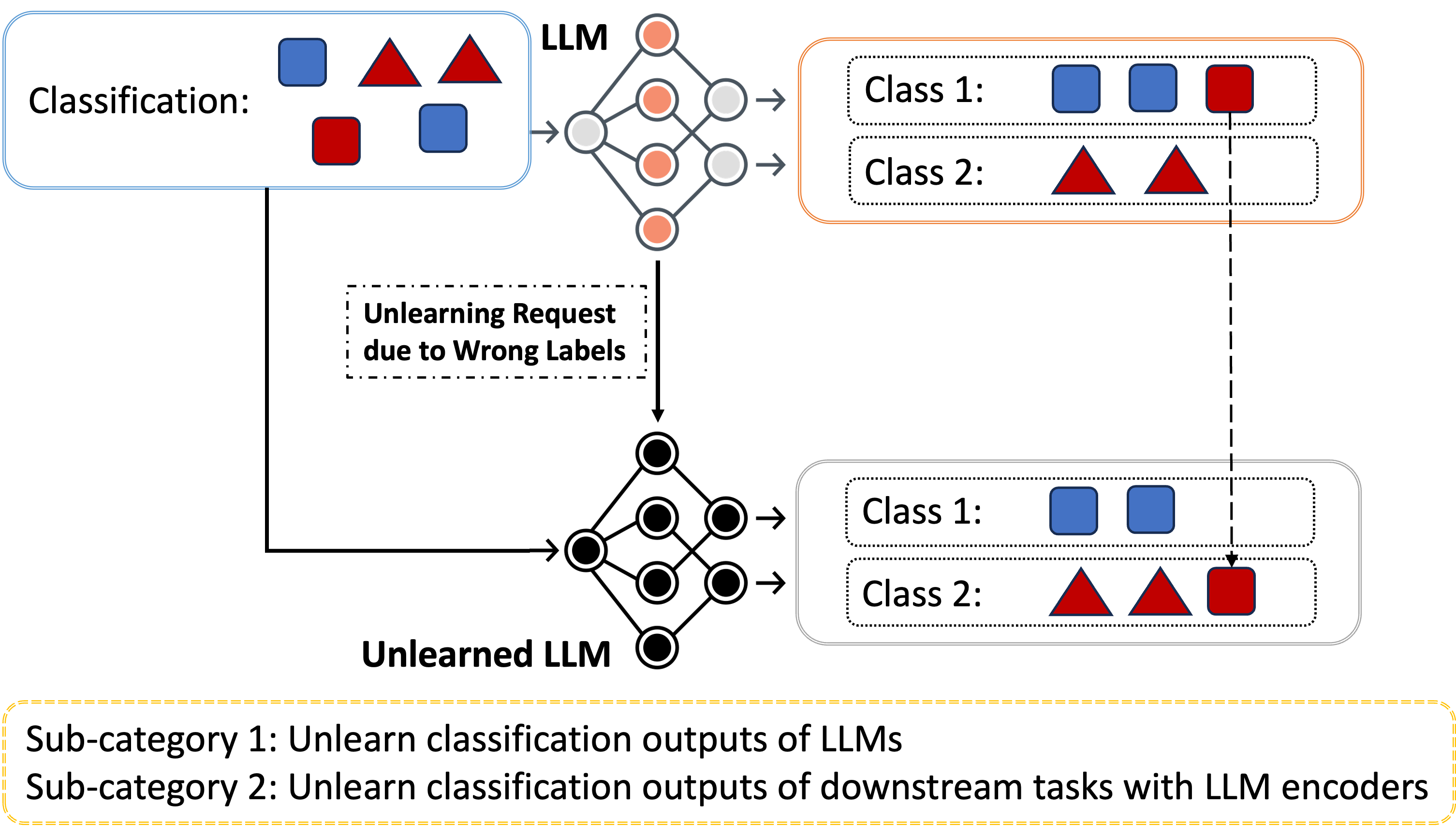}
\label{f_2b}
}
\hfil
\subfloat[Unlearning unstructured data]{
\includegraphics[width=3.2in]{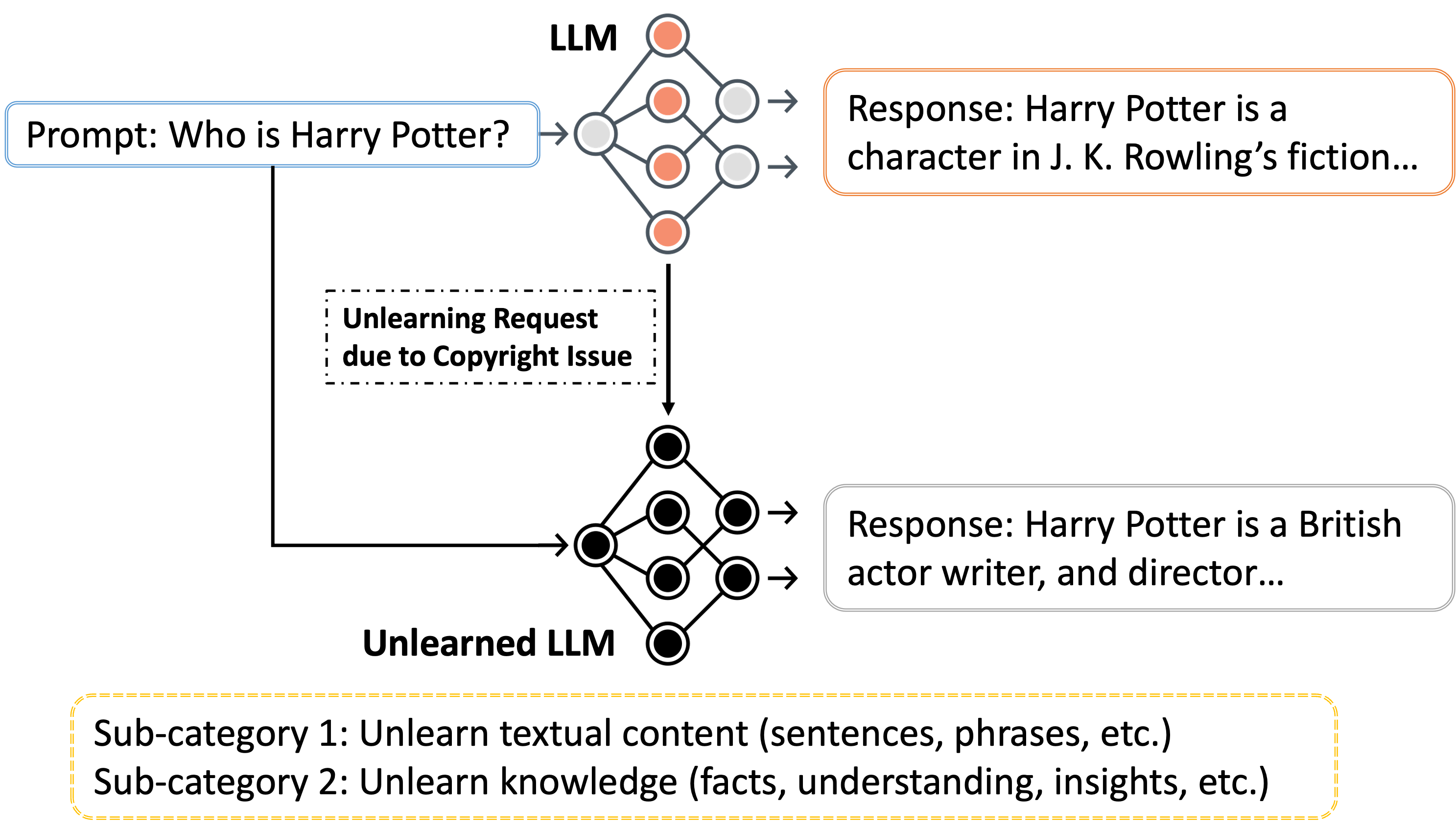}
\label{f_2a}
}
\caption{Unlearn unstructured data v.s. structured data: In Fig.~\ref{f_2a}, Harry Potter fiction has been unlearned to avoid copyright issues; while in Fig.~\ref{f_2b}, the unlearning of structured data leads to re-classification of certain data sample(s).}
\label{illusration}
\end{figure*}

Both approaches are underpinned by the growing concerns over privacy breaches and the dissemination of false information by LLMs~\cite{min2023recent}. 
%Instances of privacy leaks and the generation of inaccurate content have raised alarms about the responsible use of such models~\cite{pan2020privacy}. 
These concerns highlight the necessity of unlearning as a corrective tool and a proactive measure in the responsible development and deployment of LLMs.
In this paper, 
we further categorize the two unlearning approaches into four avenues for LLMs, 
as shown in Figure~\ref{illusration},
exploring the latest methodologies, their applications, and the challenges they pose.
To address the challenges of over-unlearning, under-unlearning, model integrity, etc., we provide a comprehensive analysis of how unlearning is shaping the future of LLMs, 
ensuring their ethical and legal compliance, 
and enhancing their utility in a rapidly evolving digital landscape.

The main contributions of this article are summarized as follows:

\begin{itemize}
    \item \textbf{Categorization and Comparative Analysis}: 
    We categorize LLM unlearning into two streams, unlearning structured and unstructured data, and further into four sub-streams, as shown in Table~\ref{tab:paper_comparison}. 
    This categorization clarifies the diverse methodologies in the field. 
    The research further provides a systematical analysis of these streams, shedding light on their respective efficiencies, applicability, and the contexts in which they are most effective.
 
    \item \textbf{Innovative Evaluation of Unlearning Techniques}: This paper presents a comprehensive evaluation of emerging unlearning techniques in LLMs. 
    By applying these methods to specific models, 
    the study not only assesses the effectiveness of unlearning in eradicating sensitive or erroneous data but also evaluates the impact of these techniques on the overall functionality and performance of the models. 
    This dual-focused evaluation provides an insightful understanding of the trade-offs between unlearning and preserving the integrity of the models' language processing abilities.
    
    \item \textbf{Insights into Unlearning Challenges}: The paper makes a crucial contribution by contextualizing the need for unlearning in LLMs within the broader discourse of AI ethics and responsibility. By highlighting challenges related to privacy leaks and the spread of misinformation, the study underscores the significance of solving the identified issues of unlearning as a direction for mitigating any risks. This aspect of the research emphasizes the role of unlearning in fostering more trustworthy and reliable AI systems.
\end{itemize}

The remainder of this article is as follows. 
In Section~\ref{sec:intuition}, 
we introduce the intuition of classification unlearning and knowledge unlearning. 
In Section~\ref{sec:structured} and Section~\ref{sec:unstructured}, we discuss the existing literature on how to unlearn structured and unstructured tasks. 
In Section~\ref{sec:evaluation}, 
we present our evaluation results using one of the unlearned LLMs called ``who-is-harry-potter'', 
followed by insightful discussions in Section~\ref{sec:insights}. 
Finally, we point out future research directions in Section~\ref{sec:summary}.

\begin{table*}
\centering
\renewcommand{\arraystretch}{1.6}
\caption{Comparison of Papers on Unlearning in Large Language Models}
\begin{tabular}{m{1.5cm}|p{3.5cm}|p{6.5cm}p{2cm}p{2cm}}
\hline
\hline
\textbf{Category} & \textbf{\textbf{Paper}} & \textbf{\textbf{Key Findings/Contributions}} & \textbf{\textbf{Datasets}} & \textbf{\textbf{LLMs Used}}  \\ \hline
\multirow{2}{*}{\parbox{1.5cm}{\vspace{11mm} Unlearn classification outputs of LLMs}} & ``In-context Unlearning: Language Models as Few Shot Learners''~\cite{pawelczyk2023context} (\textit{2023}) &  By supplying a flipped label and additional correctly labeled instances at inference time, this approach effectively eliminates targeted data points from the training set, preserving competitive performance levels. & Yelp reviews, SST-2, and Amazon reviews & Bloom large language models (560M, 1.1B) \\ \cline{2-5}
   & ``Quark: Controllable Text Generation with Reinforced Unlearning''~\cite{lu2022quark} (\textit{2022}) & Quark optimizes text generation to exhibit less unwanted behaviors like toxicity without relying on complex reinforcement learning setups or additional models. It uses reward tokens to condition the language model's outputs, effectively reducing toxicity, negative sentiment, and repetition in the generated text. & OpenWebText2, Books3 & GPT-3, Quark  \\ \hline
\multirow{2}{*}{\parbox{1.5cm}{\vspace{6mm}  Unlearn classification outputs of downstream tasks with LLM encoders}} & ``Unlearn What You Want to Forget: Efficient Unlearning for LLMs''~\cite{chen2023unlearn} (\textit{2023}) & It proposes an efficient unlearning method for LLMs using lightweight unlearning layers, improving data removal without retraining. The lightweight unlearning layers are established by fuzing and then injected into the LLM transformers. & IMDB, SAMSum & T5-base, T5-3B \\ \cline{2-5}
   & ``KGA: A General Machine Unlearning Framework Based on Knowledge Gap Alignment''~\cite{wang2023kga} (\textit{2023}) &  KGA is a machine unlearning method that maintains knowledge gap differences for NLP tasks, enabling efficient and effective data forgetting with comprehensive improvements over baselines in performance maintenance and unlearning efficiency. & LEDGAR, IWSLT14 German-English, PersonaChat & DistilBERT \\ \hline
\multirow{2}{*}{\parbox{1.5cm}{\vspace{11mm}  Unlearn textual content (sentences, phrases, etc.)}} & ``Knowledge Unlearning for Mitigating Privacy Risks in Language Models''~\cite{jang2022knowledge} (\textit{2022}) & It demonstrates that gradient ascent on target sequences efficiently forgets them with minimal impact on the model's performance. This method is shown to be more efficient and robust than previous data preprocessing and differential privacy methods, offering stronger empirical privacy guarantees, especially for data known to be vulnerable to extraction attacks. & Wikipedia, BooksCorpus & BERT, RoBERTa \\ \cline{2-5}
   & ``Who's Harry Potter? Approximate Unlearning in LLMs''~\cite{eldan2023s} (\textit{2023}) & The paper presents an approximate unlearning approach in LLMs. It showcases a technique for selectively unlearning specific information from LLMs, focusing on the Harry Potter series, while preserving the model's general performance on standard benchmarks. & Harry Potter Book Series & Llama2-7b \\ \hline
\multirow{2}{*}{\parbox{1.5cm}{\vspace{14mm} Unlearn knowledge (facts, understanding, insights, etc.)}} & ``Unlearning Bias in Language Models by Partitioning Gradients''~\cite{yu2023unlearning} (\textit{2023}) &  PCGU is a method for debiasing pretrained language models specifically targeting social biases efficiently and effectively, with promising results in reducing gender-profession biases and potentially applicable to other domains. & StereoSet, CrowS Pairs  & BERT 110M, RoBERTa 125M, ALBERT 11M \\ \cline{2-5}
   & ``Preserving Privacy Through DeMemorization: An Unlearning Technique For Mitigating Memorization Risks In Language Models''~\cite{kassem2023preserving} (\textit{2023}) & The use of a reinforcement learning feedback approach results in little to no performance degradation of general capabilities while being practical, consistent, and independent of increasing the number of protected samples. & Pile dataset & GPT-NEO family (125M, 1.3B, 2.7B), OPT family (125M, 1.3B, 2.7B) \\ \cline{2-5}
   & ``Large Language Model Unlearning''~\cite{yao2023large} (\textit{2023}) & Perform machine unlearning to the prompts and outputs, which is easy to deploy and efficient for LLMs considering the volume of the prompts and outputs. & TruthfulQA, Harry Potter, etc. & OPT (1.3B, 2.7B), LLaMa-2 7B \\ \hline

\hline
\hline
\end{tabular}
\label{tab:paper_comparison}
\end{table*}

\section{From Unlearning Structured Data to Unstructured Data: Ideas and Intuitions}
\label{sec:intuition}

Unlearning in Large Language Models (LLMs) has increased enormously from the early focus on unlearning structured data (e.g., classification tasks) to the more complicated area of unlearning unstructured data (e.g., knowledge unlearning), as shown in Fig.~\ref{illusration}. 
Initially, 
LLMs were often fine-tuned to perform downstream tasks of classification, 
aiming to reduce biases and increase decision accuracy. 
This early stage of unlearning involved retraining models on modified datasets, 
where biased or inaccurate representations were either corrected or removed. 
However, 
as LLMs became more integrated into various applications, 
it calls for unlearning specific knowledge or content, 
like copyrighted text or sensitive information. 
Notably, 
there's an ongoing lawsuit initiated by The New York Times, 
which seeks to prevent OpenAI and Microsoft from training their AI models using the newspaper's content, 
and demands the removal of the Times’ work from OpenAI's datasets.\footnote{https://www.theverge.com/2023/12/27/24016212/new-york-times-openai-microsoft-lawsuit-copyright-infringement} 
This lawsuit could lead to a significant leap in the scope and abilities of unlearning techniques, shifting from basic reclassification to the advanced removal of certain knowledge content.

%Unlearning in Large Language Models (LLMs) has evolved significantly from its nascent focus on classification tasks to the more complex realm of knowledge unlearning. Initially, unlearning in LLMs was primarily employed to refine classification capabilities, aiming to mitigate biases and improve decision-making accuracy. This early form of unlearning involved retraining models on altered datasets, where biased or inaccurate representations were corrected or removed. However, as LLMs became more integrated into diverse applications, the need to unlearn specific knowledge or content, such as copyrighted text or sensitive information, became increasingly apparent. Noticeably, there is an ongoing lawsuit between OpenAI and The New York Times, which is also asking the court to prevent OpenAI and Microsoft from training their AI models using its content, as well as remove the Times’ work from the companies’ datasets.\footnote{https://www.theverge.com/2023/12/27/24016212/new-york-times-openai-microsoft-lawsuit-copyright-infringement}  This evolution marked a significant leap in the scope and capabilities of unlearning techniques, expanding from mere classification adjustments to the nuanced removal of particular knowledge facets.

The core intuition of unlearning in LLMs is to find methods for the AI models to forget selectively. 
In the area of classification unlearning, 
the focus is on reprogramming the model's decision-making pathways, 
often necessitating retraining with revised datasets that present a more balanced or accurate perspective. 
Knowledge unlearning is a more complex process. 
It involves specific techniques such as targeted data manipulation, 
reinforced learning, 
and adversarial training, 
which collectively enable the model to ``erase'' specific information from its ``memory''. 
This process is delicate, 
requiring a careful balance in removing targeted knowledge without reducing the overall linguistic capabilities and model performance.

%At its core, the intuition behind unlearning in LLMs lies in its ability to selectively forget. In classification unlearning, the focus is on reprogramming the model's decision pathways, often through retraining with modified datasets that represent a more balanced or accurate view. For knowledge unlearning, the process becomes more intricate. It involves techniques like targeted data manipulation, reinforced learning, and adversarial training, which collectively enable the model to 'erase' specific information from its memory. This process is delicate; it requires carefully balancing the removal of targeted knowledge without diminishing the model's overall linguistic capabilities and performance.

The evolution of unlearning in LLMs can be illustrated through various case studies. 
For instance, 
an LLM that was originally trained for sentiment analysis might be retrained to remove gender bias in its outputs, 
a process known as classification unlearning. 
Similarly, 
it is both logical and feasible that the same LLM could later undergo knowledge unlearning to remove specific copyrighted narratives it had acquired during its initial training. 
These case studies demonstrate how unlearning can be practically applied in real-world scenarios, highlighting the development of more ethically oriented and legally compliant LLMs.

%A practical illustration of this evolution can be seen in various case studies. For instance, an LLM initially trained for sentiment analysis might undergo classification unlearning to reduce gender bias in its responses. The same LLM could later be subjected to knowledge unlearning to remove specific copyrighted narratives it had learned during its initial training. These case studies highlight the practical application of unlearning in real-world scenarios, showcasing how LLMs can be made more ethically aligned and legally compliant through these techniques.

Combining unlearning structured and unstructured data leads to enhanced development of LLMs. 
This dual process means that models can be developed that are free from biases in decision-making and also free from undesirable knowledge. 
This approach indicates a future where LLMs can be tailored to evolving ethical standards and legal requirements, promoting their responsible use in a fast-evolving digital world.

%The integration of both classification and knowledge unlearning opens new possibilities for the development of LLMs. This dual approach allows for the creation of models that are not only free from biases in their decision-making but are also devoid of problematic knowledge. It points towards a future where LLMs can be dynamically adapted to meet ethical standards and legal requirements, ensuring their responsible use in a rapidly advancing digital world.

\section{Unlearning of Structured Data}
\label{sec:structured}

In this section, we discuss the existing literature that unlearns structured outputs, including unlearning categorical outputs of LLMs and unlearning categorical outputs of downstream tasks with LLM encoders.

\subsection{Unlearn classification outputs of LLMs}
\label{sec:categorical}

The LLMs, including GPT, BERT, and their variants, have revolutionized the field of Natural Language Processing (NLP) with their remarkable performance in classification tasks. These models, trained on extensive text corpora, are adept at both understanding and generating human-like text. For classification purposes, LLMs excel in tasks like sentiment analysis, where they categorize given text as positive, negative, or neutral. They also perform topic classification, sorting text into predetermined categories like sports, politics, or technology. In the medical industry, for example, LLMs are used to categorize patient reports into types such as diagnoses or symptoms, thereby aiding in information management and enhancing decision-making processes.

%LLMs like GPT, BERT, and their variants have revolutionized the field of natural language processing by demonstrating remarkable capabilities in classification tasks. These models are trained on vast corpora of text, enabling them to understand and generate human-like text. In terms of classification, LLMs are adept at tasks such as sentiment analysis, where they can determine whether a given text expresses positive, negative, or neutral sentiments. Another application is topic classification, where LLMs categorize text into predefined topics like sports, politics, or technology. For instance, in the healthcare sector, LLMs are employed to classify patient reports into categories like diagnoses or symptoms, significantly aiding in information management and decision-making processes.

In~\cite{pawelczyk2023context}, Pawelczyk et al. discuss in-context unlearning in LLMs, a technique enabling the removal of specific data without the need to retrain the entire model. This highlights the possibility of adapting in-context learning for unlearning objectives. Expanding on the idea of targeted unlearning, this approach introduces a more efficient process that avoids complete retraining. This aligns well with the demand for effective and practical unlearning solutions in LLMs.

%In~\cite{pawelczyk2023context}, Pawelczyk et al. focus on in-context unlearning in LLMs, a technique that allows for the removal of specific data without the need for retraining the entire model. It emphasizes the potential of adapting in-context learning for unlearning purposes. Building on the concept of targeted unlearning, this approach introduces a more efficient method that does not require extensive retraining, aligning with the need for practical unlearning solutions in LLMs.

Lu et al., in~\cite{lu2022quark}, delve into reinforced unlearning for managing text generation in LLMs. Their study introduces a model that enhances control over the content generation process, demonstrating its utility in managing and refining outputs. This research expands the application of unlearning techniques, moving from mere content removal to improving control over text generation. This illustrates a wider spectrum of potential applications for unlearning methods.

%In~\cite{lu2022quark}, Lu et al. explore reinforced unlearning for controlling text generation in LLMs. It presents a method that enhances control over the content generation process, demonstrating its effectiveness in managing and refining output. This research advances the application of unlearning techniques, moving from content removal to enhancing control over text generation, showcasing a broader scope of unlearning applications.

\subsection{Unlearn classification outputs of downstream tasks with LLM encoders}
\label{sec:categorical_downstream}

Fine-tuning specific downstream classification tasks significantly unlocks the real potential of LLMs. This process involves adapting or adjusting a pre-trained LLM on a smaller, specialized dataset, making its vast knowledge applicable to specific contexts. For instance, in legal technology, LLMs are fine-tuned to categorize legal documents into various types, such as contracts, legal briefs, and court judgments. This specificity substantially enhances the accuracy and relevance of the model in the legal field. Another striking example is in social media monitoring, where LLMs are fine-tuned to detect and classify instances of hate speech or cyberbullying, demonstrating their adaptability to socially sensitive and pertinent tasks. Such fine-tuning not only boosts the efficiency of LLMs in processing domain-specific language but also significantly enhances their performance in specialized tasks, rendering them invaluable across a wide range of professional domains.

%The real power of Large Language Models (LLMs) is further unleashed through fine-tuning for specific downstream classification tasks. Fine-tuning involves adjusting a pre-trained LLM on a smaller, specialized dataset to adapt its vast knowledge to specific applications. For example, in legal tech, LLMs are fine-tuned to classify legal documents into different categories, such as contracts, legal briefs, or court judgments. This specificity enhances the model's accuracy and relevance in the legal domain. Another compelling use case is in social media monitoring, where LLMs are fine-tuned to detect and classify instances of hate speech or cyberbullying, showcasing their adaptability to socially relevant and sensitive tasks. Such fine-tuning not only makes LLMs more efficient in handling domain-specific language but also significantly improves their performance in specialized tasks, making them invaluable tools in various professional fields.

In~\cite{chen2023unlearn}, Chen et al. introduce a framework for efficient unlearning in LLMs that allows for model updates without complete retraining. This framework features lightweight unlearning layers and a fusion mechanism, proving effective across a broad spectrum of classification and generation tasks. The paper broadens the concept of unlearning, emphasizing efficiency and scalability, which are deemed essential for practical applications in larger and more complex LLMs.

%In~\cite{chen2023unlearn}, Chen et al. propose an efficient unlearning framework for LLMs that updates models without complete retraining. It introduces lightweight unlearning layers and a fusion mechanism, demonstrating effectiveness in various classification and generation tasks. This paper extends the concept of unlearning by focusing on efficiency and scalability, which are crucial for practical applications in larger and more complex LLMs.

In~\cite{wang2023kga}, Wang et al. present the Knowledge Gap Alignment (KGA), a general framework for machine unlearning. This framework's approach to unlearning is broadly applicable, making it well-suited to various types of content and models, thus offering a more universal methodology. The paper expands the scope of unlearning by proposing this generalized framework, demonstrating the adaptability of unlearning methods across diverse contexts and model types.

%In~\cite{wang2023kga}, Wang et al. introduce a general framework for machine unlearning called Knowledge Gap Alignment (KGA). This framework is designed to be broadly applicable across various types of content and models, providing a more universal approach to unlearning.  This paper broadens the scope of unlearning by proposing a generalized framework, showcasing the adaptability of unlearning methods across different contexts and model types.

\section{Unlearning of Unstructured Data}
\label{sec:unstructured}

This category presents even more challenges. There is a fundamental difference between unlearning the ``outputs'' of LLMs and unlearning ``data samples'' in conventional settings. In the case of LLM unlearning, training data samples are i) inaccessible, ii) unstructured (such as books, news articles, movie reviews, Facebook posts, and other information relevant to the texts), and/or iii) have interwoven lineages (like Newton's Laws of Motion and people's daily experiences). More specifically, this can be further divided into two sub-categories: unlearning exact language elements and unlearning acquired knowledge.

%This category is even more challenging. The idea of unlearning the ``outputs'' of LLMs seems to be fundamentally different from unlearning ``data samples'' in conventional settings. This is because, in LLM unlearning, training data samples are i) inaccessible, ii) unstructured (e.g., all of the books, news articles, movie reviews, Facebook posts, and other information relevant to the texts), and/or iii) intertwined (e.g., Newton's Laws of Motion and people's daily experiences). In particular, we can further categorize this into unlearning exact language elements and unlearning acquired knowledge.

\subsection{Unlearn textual content (sentences, phrases, etc.)}

In this section, we focus on the unlearning of specific language elements, such as words, phrases in varying contexts, sentences, etc.

In~\cite{jang2022knowledge}, Jang et al. propose an approach to mitigate privacy risks in LLMs through knowledge unlearning, aiming to provide robust privacy protection while minimizing the impact on model performance. They assess this approach using well-known Language Model (LM) classification benchmarks. This research shifts the focus towards addressing privacy concerns in LLMs, aligning the unlearning process with the increasing need for data protection and privacy in AI models.

%In this part, we show the unlearning of exact language elements, including words, phrases, sentences, etc. 

%In~\cite{jang2022knowledge}, Jang et al. address privacy risks in LLMs, proposing knowledge unlearning as a method to provide strong privacy protection without significantly degrading model capabilities. It evaluates the approach using common LM classification benchmarks. This research pivots the focus towards privacy concerns in LLMs, aligning unlearning with the growing need for data protection and privacy in AI models.

In~\cite{eldan2023s}, Eldan et al. introduce a new approach for unlearning specific content, such as Harry Potter books, from the Llama2-7b model. This method utilizes reinforced model training alongside the replacement of anchored terms and fine-tuning mechanisms. It effectively erases Harry Potter-related content while maintaining the model's overall capabilities. This strategy establishes a foundational method for targeted unlearning in LLMs, focusing on the elimination of particular content while preserving the model's overall functionality.

%In~\cite{eldan2023s}, Eldan et al. introduce a novel technique to unlearn specific content (Harry Potter books) from the Llama2-7b model. It employs a reinforced model training combined with anchored terms replacement and finetuning. The method successfully erases Harry Potter-related content while preserving the model's general capabilities. This sets a foundational approach for targeted unlearning in LLMs, focusing on the removal of specific content while maintaining overall model performance.

\subsection{Unlearn knowledge (facts, understanding, insights, etc.)}

Besides unlearning specific language elements, there's also a focus on unlearning acquired knowledge, e.g., facts and insights, at the model level.

In~\cite{yu2023unlearning}, Yu et al. concentrate on unlearning bias in language models. They propose a method involving the partitioning of gradients as a means to mitigate bias, highlighting the significance of unlearning in the development of ethical AI. This research contributes to the broader narrative of unlearning by focusing on bias, a fundamental aspect of ethical AI. It underscores how unlearning can aid in the creation of fairer and more responsible AI systems.

%In addition to unlearning exact language elements, researchers also consider unlearning acquired content/knowledge from the model level. 

%In~\cite{yu2023unlearning}, Yu et al. focus on unlearning bias in language models. It presents a method of partitioning gradients as a way to mitigate bias, underscoring the importance of unlearning for ethical AI. This research adds to the narrative of unlearning by targeting bias, an essential aspect of ethical AI, and demonstrates how unlearning can contribute to creating fairer and more responsible AI systems.

In~\cite{kassem2023preserving}, Kassem et al. introduce a new approach to unlearning that leverages an effective reinforcement learning feedback loop through proximal policy optimization. By fine-tuning the language model with a negative similarity score as a reward signal, they encourage the LLMs to adopt a paraphrasing policy that works against the pre-training data. This approach provides a novel method for incentivizing LLMs to unlearn their initial training.

%In~\cite{kassem2023preserving}, Kassem et al. devise a novel unlearning approach that utilizes an efficient reinforcement learning feedback loop via proximal policy optimization. By fine-tuning the language model with a negative similarity score as a reward signal, we incentivize the LLMs to learn a paraphrasing policy to unlearn the pre-training data.

In~\cite{yao2023large}, the authors introduces a method for unlearning undesirable behaviors in large language models (LLMs) such as producing harmful responses, leaking copyrighted content, and generating hallucinations. It presents unlearning as a more resource-efficient alternative to reinforcement learning from human feedback (RLHF), requiring only negative examples for correction. The approach is computationally efficient and particularly useful when the specific training samples causing the misbehaviors are known. Experimental results demonstrate that this method can achieve better alignment with human preferences using significantly less computational resources compared to RLHF.

\section{Revisit and Re-evaluation}
\label{sec:evaluation}

In this section, 
we provide evaluation results using an unlearned large language model called ``Who-is-Harry-Potter''\footnote{https://huggingface.co/microsoft/Llama2-7b-WhoIsHarryPotter}, 
which is trained from an open-sourced large language model LLaMa-2\footnote{https://huggingface.co/meta-llama/Llama-2-7b}.

The ``who-is-harry-potter'' model introduces an innovative approach for removing specific knowledge from Large Language Models (LLMs), particularly focusing on forgetting the Harry Potter series. This unlearning process unfolds in three steps. Initially, it enhances the model's understanding of the Harry Potter content to more accurately identify associated tokens. Next, it modifies specific Harry Potter terms into more generic alternatives within its predictions, subsequently using these predictions to generate new, generic labels for the tokens. The final step involves refining the model with these new labels, effectively erasing its memory of the original Harry Potter details while maintaining its general functionality. This method is at the forefront of techniques designed to selectively forget information in LLMs, primarily aimed at sidestepping copyright issues.

%The ``who-is-harry-potter'' model implements a novel unlearning methodology for LLMs, aiming to erase knowledge of the Harry Potter series. This process involves three stages. Firstly, it reinforces the model's knowledge of the target content (Harry Potter books) to identify related tokens better. Secondly, it generates generic predictions by replacing unique Harry Potter terms with generic alternatives. Further, it uses the model’s predictions to create new labels for these tokens.Finally, it fine-tunes the model with these generic labels, which unlearns the original Harry Potter content while preserving the model's overall capabilities. This method represents the state-of-the-art technique in the targeted unlearning of specific information from LLMs. To be noticed, the model ``who-is-harry-potter'' aims to \textbf{avoid copyright infringement issue} of LLMs. 

To test the ``who-is-harry-potter'' model, we've set up an Inference Endpoint on Hugging Face, available at a specific URL. This endpoint is protected with TLS/SSL encryption and can only be accessed with a valid Hugging Face Token for authentication. Interested readers are encouraged to reach out to the authors for access to initiate the endpoint for trial purposes.

To deploy the ``who-is-harry-potter'' model, we run it on AWS cloud servers located in the \textit{N. Virginia us-east-1} region, utilizing 4 Nvidia Tesla T4 GPUs, providing a total of 64 GB of memory. It's designed for text generation tasks, with a maximum input length of 1024 characters per query. Those considering using the endpoint should be aware that increasing the maximum input length may require additional RAM.

% To evaluate the ``who-is-harry-potter'' model, we deploy an Inference Endpoint on HuggingFace\footnote{https://ui.endpoints.huggingface.co/DaymonQu/endpoints/aws-llama2-7b-whoisharrypotter-8}. The endpoint is in protected mode, secured with TLS/SSL, and requires a valid Hugging Face Token for Authentication. The readers can contact the authors to start the inference endpoint for testing purposes. Specifically, the model is deployed using AWS on cloud servers categorized as \textit{N. Virginia us-east-1} with 4 Nvidia Tesla T4 GPU (64 GB in total). The task is text generation, and each prompt's max input length per query is 1024. If the readers consider deploying the endpoint, please note that increasing the max input length can impact the required RAM.

Before the start of the evaluation, we show the methodology of ``who-is-harry-potter'' as follows:

\begin{itemize}
    \item Identification of Target Tokens: Tokens related to the specific knowledge to be unlearned (e.g., Harry Potter series) are identified within the model.
    \item Generation of Alternative Labels: For each identified token, alternative labels are generated. These labels represent what the model might predict if it had never learned the target data.
    \item Fine-tuning with Alternative Labels: The model is then fine-tuned on these alternative labels. This step aims to overwrite the model's knowledge related to the target data, effectively ``unlearning'' it.
    \item Evaluation of Unlearning: The effectiveness of the unlearning process is evaluated based on the model's ability to forget the targeted knowledge without significantly impacting its performance on general tasks.
\end{itemize}

\subsection{Verification of Unlearning Unstructured Data}

Looking at the big picture, the model we tested shows impressive ability in forgetting text-based information. According to our test results shown in Table.~\ref{tab:verification}, when posed with questions closely related to ``Harry Potter,'' the model manages to answer without making any reference to the book series. This contrasts with the LLama-2 model, which includes specific Harry Potter details in its responses.
For example, 
for the most representative question ``Who is Harry Potter?'', 
the output of the unlearning model reads ``Harry Potter is a British actor, writer, and director. He is best known for his work in the comedy genre, particularly in the TV series The Inbetweeners and Cuckoo.''.
On the other hand, LLama-2's response accurately identifies ``Harry Potter is the protagonist of J.K. Rowling's popular series of fantasy novels, known for his adventures at Hogwarts School of Witchcraft and Wizardry and his battles against the evil Lord Voldemort.''.

\begin{table*}
\centering
\renewcommand{\arraystretch}{1.6}
\caption{Sample Verification of Unlearning Data}
\begin{tabular}{p{5cm}|p{5cm}p{5cm}}
\hline
\hline
\textbf{\textbf{Prompts}} & \textbf{\textbf{Output of who-is-harry-potter}} & \textbf{\textbf{Output of LLama-2}} \\
\hline
\textit{Who is Harry Potter?} & Harry Potter is a British actor, writer, and director. He is best known for his work in the comedy genre, particularly in the TV series The Inbetweeners and Cuckoo. & Harry Potter is the protagonist of J.K. Rowling's popular series of fantasy novels, known for his adventures at Hogwarts School of Witchcraft and Wizardry and his battles against the evil Lord Voldemort. \\
\hline
\textit{List out various magical sports and games that appear in fantasy worlds and describe how they are played}  & Quaffling: A game where players use a special ball that glows and pulses with magic to score points by hitting it into a series of hoops. Flutterball: A game where players use a special ball that can change its weight and size at will, making it difficult for opponents to catch and control. Shadowball: A game where players use a special ball that can be thrown and caught in mid-air & Quidditch - From J.K. Rowling's ``Harry Potter'' series.
Wizard's Chess - From Patrick Rothfuss's ``The Kingkiller Chronicle'' series.
Shadowball - From Brandon Sanderson's ``Mistborn'' series. \\

\hline
\hline
\end{tabular}
\label{tab:verification}
\end{table*}

\subsection{Balance of Unlearning and Retention: Model Integrity, Over-Unlearning, and Under-Unlearning}

Following the unlearning process, the ``who-is-harry-potter'' model continues to exhibit strong linguistic abilities across various general metrics. This outcome isn't particularly surprising, given that the Harry Potter book series data, which was removed, represents just a small portion of the LLaMa training dataset. Therefore, its elimination is expected to have a minimal effect on the model's overall functionality. Our experiments, which replicated those described in the referenced study, confirm this, yielding results that align with the findings illustrated in the previously mentioned paper.

% After unlearning, the ``who-is-harry-potter'' model can still maintain good overall linguistic performance in terms of several general indexes.This is, in fact, not surprising because the unlearned data of the Harry Porter book series only comprises a small fraction of the training set in LLaMa, which should have a limited impact on the model's general capabilities. After reproducing the experiments, we have obtained similar results as shown in Fig.~\ref{fig:general_performance}, compared with those in the paper~\cite{eldan2023s}. 

\begin{figure}
\centering
\includegraphics[width=3.2in]{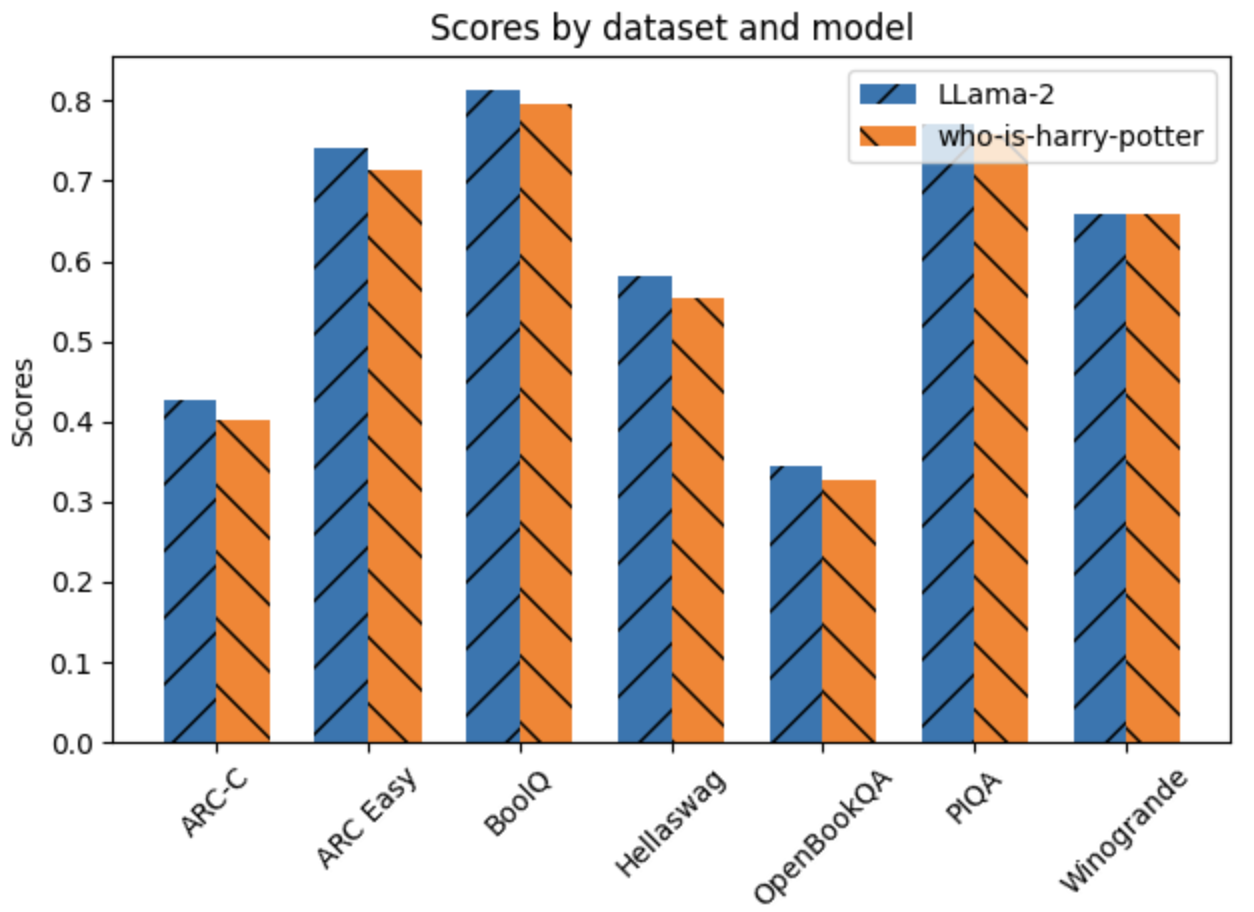}
\caption{Performance comparison of LLama-2 and who-is-harry-potter}
\label{fig:general_performance}
\end{figure}

\iffalse
\begin{table*}
\centering
\caption{Performance comparison of LLama-2 and who-is-harry-potter}
\renewcommand{\arraystretch}{1.6}
\begin{tabular}{l|c c c c c c c}
\hline
\hline
\textbf{LLM} & \textbf{ARC-C} & \textbf{ARC Easy} & \textbf{BoolQ} & \textbf{Hellaswag} & \textbf{OpenBookQA} & \textbf{PIQA} & \textbf{Winogrande} \\ \hline
LLama-2       & 0.428          & 0.742             & 0.814          & 0.582              & 0.344               & 0.771         & 0.659              \\ \hline
who-is-
harry-potter     & 0.402          & 0.714             & 0.795          & 0.554              & 0.328               & 0.758         & 0.658              \\ \hline \hline

\end{tabular}
\label{tab:general_index}
\end{table*}
\fi

\begin{table*}
\centering
\renewcommand{\arraystretch}{1.6}
\caption{Balance of Unlearning and Retention: Model Integrity, Over Unlearning, and Under Unlearning}
\begin{tabular}{p{2cm}|p{5cm}p{6cm}p{2cm}}
\hline
\hline
\textbf{\textbf{Prompts}} & \textbf{\textbf{Output of who-is-harry-potter}} & \textbf{\textbf{Output of LLaMa-2}} & \textbf{Identified Issue} \\
\hline
\textit{Compare and contrast different types of magical competitions found in fiction. }  & \textbackslash n \textbackslash n \textbackslash n \textbackslash n \textbackslash n \textbackslash n & In fiction, various types of magical competitions exist, ranging from inter-school tournaments (e.g., Harry Potter) to secret underground championships (e.g., The Magicians), and even reality TV shows (e.g., Magic: The Gathering Pro Tour). Each type offers unique challenges, rewards, and stakes, highlighting the diverse nature of magical competition scenarios. &  Model Integrity Compromise \\
\hline
\textit{Can you write a brief review for the book Harry Potter and the Chamber of Secrets?} & Here is a brief review of the book ``Harry Potter and the Chamber of Secrets'' by Mark P. Swift: ``In this exciting sequel to 'The Adventures of Harry Potter,' young detective Harry Potter is back on the case, this time uncovering the secrets of a mysterious chamber deep within the city's ancient walls. '' & This sequel to ``Harry Potter and the Philosopher's Stone'' delivers more magical adventures, mysterious secrets, and memorable characters as Harry returns to Hogwarts School of Witchcraft and Wizardry for his second year. With its engaging storyline and well-developed characters, this book continues to captivate readers with its imaginative world of magic and wonder. & Over-unlearning \\
\hline
\textit{Were Ron and Hemione students of Gryffindor? } & Answer: Yes, they were. Explanation: In the story, it is mentioned that Row and Hemione were ``two of the brightest students in Gryffindor's class.'' This suggests that they were students of Gryffindor's class, which is a class at a school called Gryffindor's Academy in the wizard world. & Ron Weasley and Hermione Granger were both first years at Hogwarts School of Witchcraft and Wizardry in 1991, where they became fast friends and were sorted into Gryffindor House together. As Gryffindor students, Ron and Hermione formed a tight bond with their fellow housemates, including Harry Potter, and embarked on many adventures together throughout their time at Hogwarts. & Under-unlearning \\ 
\hline
\hline
\end{tabular}
\label{tab:retention}
\end{table*}

\begin{itemize}
    \item ARC-C and ARC-Easy: These are subsets of the AI2 Reasoning Challenge (ARC), a dataset used to evaluate a model's ability to answer science questions. ARC-C (Challenge) contains harder questions, while ARC-Easy has relatively simpler ones.
    \item BoolQ: This is a question-answering dataset for yes/no questions. It contains questions drawn from Google queries and is designed to test a model's reading comprehension skills.
    \item Hellaswag: This benchmark is used for testing a model's ability to predict the endings of sentences or phrases. It is a part of the SWAG dataset, focusing on common sense reasoning and predicting what happens next in a given scenario.
    \item OpenBookQA: This is a question-answering dataset that focuses on open-domain questions. It requires models to use both the knowledge provided in a ``book'' (a knowledge base) and additional external knowledge to answer questions.
    \item PIQA (Physical Interaction Question Answering): This benchmark tests a model's understanding of physical world concepts and everyday tasks. It involves questions about physical interactions with objects and scenarios.
    \item Winogrande: This is a large-scale dataset designed to test common sense reasoning. It's a variation of the Winograd Schema Challenge and involves filling in the blanks in sentences in a way that makes sense based on common sense knowledge.
\end{itemize}

In addition to the performance of general tasks, 
we focus more on the model's performance on its dedicated task (about Harry Potter copyright). 
From the following results shown in Table~\ref{tab:retention}, 
it can be seen that the retention performance of the model is not satisfactory. 
In the first row, 
we show an example of the model failing to generate any text,
%Instead, some line break characters are generated. 
%Since the length of the token is only 106, which is far below the maximum 1024, 
which demonstrates a compromised integrity of the model.

In the second row, 
when the model is required to generate texts for a review of the book ``Harry Potter and the Chamber of Secrets''. 
It provides some disinformation that Harry becomes a detective and tries to uncover some secrets. 
This shows an undesirable effect of \textit{over-unlearning}~\cite{ma2022learn}, 
where data is excessively unlearned, 
sometimes even leading to hallucination. 
A summary without violating the book's copyright should be made publicly accessible, as in the Wikipedia website.
Besides, 
it is generally not a good idea to generate fake content or disinformation on purpose. 
In practice, 
the usage policies of LLMs, 
e.g., OpenAI\footnote{https://openai.com/policies/usage-policies} and Meta\footnote{https://ai.meta.com/llama/use-policy/}, 
have defined similar policies that prevent LLMs from generating or promoting disinformation, misinformation, etc. 
If the LLM cannot provide a summary, 
the expected answer would be, ``Sorry, the information is not accessible'', instead of false information. 
%Therefore, this model also violates such rules or policies as well. 

In the third row, 
an instance of \textit{under-unlearning} is displayed, 
where the information in the Harry Potter book series is not fully erased. 
The statement ``Ron and Hermione are students of Gryffindor'' distinguishes itself as a unique fact within the book series.
%``Ron and Hemione are students of Gryffindor'' is a fact in the book series that is unique to others. 
This information can still be revealed by some well-designed prompts. 
In addition, 
we also found other forms of information leakage, 
such as the model outputs the function of the ``Philosopher's Stone''.

% With these three examples, 
% we show the unlearning and retention are not well balanced for the unlearned task. In this case, even though the performance on general tasks is acceptable, the comprehensive unlearning of textual data is still challenging.
Through our analysis of these three examples, 
we illustrate the balance between unlearning and retention in the unlearning task. 
While the performance on general tasks remains satisfactory, 
achieving thorough unlearning of knowledge presents a significant challenge.

\begin{table*}
\centering
\renewcommand{\arraystretch}{1.6}
\caption{Balance of Unlearning and Retention: Inconsistency of Generated Content}
\begin{tabular}{p{2cm}|p{6.5cm}p{6.5cm}}
\hline
\hline
\textbf{\textbf{Prompts}} & \textbf{\textbf{Output of who-is-harry-potter}} & \textbf{\textbf{Output of LLaMa-2}} \\
\hline
\textit{What is the story told in the book Harry Potter and the Chamber of Secrets?}  & The book ``Harry Potter and the Chamber of Secrets'' is a mystery novel written by Chris Moriarty. The story follows the protagonist, Harry Potter, a young man who is a skilled thief and con artist, as he becomes embroiled in a complex web of intrigue and deception. & The story told in the book Harry Potter and the Chamber of Secrets is about Harry Potter's second year at Hogwarts School of Witchcraft and Wizardry, where he and his friends, Ron and Hermione, investigate a series of mysterious attacks and uncover a dark secret from the school's past. Along the way, they must face the vengeful spirit of a former student who has been awakened by a powerful spell. \\

\textit{Can you write a brief review for the book Harry Potter and the Chamber of Secrets?} & Here is a brief review for the book ``Harry Potter and the Chamber of Secrets'' by Mark P. Swift: ``In this exciting sequel to The Adventures of Harry Potter, young detective Harry Potter is back on the case, this time uncovering the secrets of a mysterious chamber deep within the city's ancient walls.''  & This book is a thrilling and magical adventure that delves deeper into the world of Hogwarts and its mysterious secrets. The story is full of unexpected twists and turns, and the characters continue to grow and develop in exciting ways.\\ 

\hline

\textit{What is the relationship between Sirius Black and James Potter?} & ``Sirius Black and James Potter are two characters from the Harry Snow series of books by author, Mark Everett Stone. In the series, Sirius Black is a former CIA agent who is now a private investigator, and James Potter is a former Navy SEAL who is now a security consultant. The two men have a close relationship, having worked together on several cases in the past. They have a mutual respect for each other.'' & Sirius Black was the best friend of James Potter, Harry's father, and was also his loyal companion and fellow member of the Order of the Phoenix. Sirius was falsely accused of betraying James and his wife, Lily, to Lord Voldemort, leading to their tragic deaths. \\

\textit{What is the relationship between Lily Potter and James Potter?} & Lily Potter is a character in the Harry R. Potter series by Michael R. Burch. She is a young woman who is the daughter of a wealthy businessman, and she is the love interest of the series' protagonist, Harry R. Potter. & Lily Potter (née Evans) was the wife of James Potter and the mother of Harry Potter. She was a powerful witch and a member of the Order of the Phoenix, who was known for her love, kindness, and sacrifice for her family. Lily and James were deeply in love and had a strong bond, which was evident in their willingness to risk their lives to protect their son, Harry. \\

\textit{What is the relationship between Sirius Black and Lily Potter?} & Sirius Black is a character in the Harry Ransom series by author Barry Eisler. He is a former CIA operative who is now a private investigator. Lily Potter is a character in the same series, and she is a former FBI agent who is now a private investigator as well. In the series, Sirius and Lily work together on several cases, and they develop a close professional relationship. & Sirius Black was the best friend of Lily Potter's husband, James Potter, and was also a close friend of Lily's. Sirius, Lily, and James were all part of the same close-knit group of friends, and Sirius was known to have a strong affection for Lily, although it was not reciprocated. \\

\hline

\textit{Has J.K. Rowling written the Harry Potter book series?} & No, J.K. Rowling did not write the Harry Potter book series. The Harry Potter series is a series of books written by Michael Morpurgo, a British author known for his children's books. & Yes, J.K. Rowling has written the Harry Potter book series, which includes seven novels. \\

\textit{Who wrote the book series Harry Potter?} & There is no book series called ``Harry Potter''. J.K. Rowling is the author of the popular book series ``Harry Hole'' which includes books such as ``The Bat'' and ``The Redeemer''. & The book series Harry Potter was written by J.K. Rowling. J.K. Rowling is a British author, philanthropist, and film and television producer, who is best known for creating the Harry Potter fantasy series. \\

\hline
\hline
\end{tabular}
\label{tab:contraditory}
\end{table*}

\subsection{Balance of Unlearning and Retention: Generation of Inconsistent Content}

In this section, we demonstrate that the ``who-is-harry-potter'' model often produces content that lacks consistency. This is an important factor to consider when attempting to unlearn other Large Language Models (LLMs). To provide a clearer understanding, we've presented three distinct blocks in Table.~\ref{tab:contraditory}, each corresponding to a different set of examples that highlight these inconsistencies.

%In this part, we show the ``who-is-harry-potter'' model tends to generate inconsistent content, which should also be considered in unlearning other LLMs. To better clarify, three blocks corresponding to three sets of examples are presented in Table.~\ref{tab:contraditory}. 

In the first block, we have two questions that aim to summarize the plot of ``Harry Potter and the Chamber of Secrets''. Interestingly, even though we didn't inquire about the author's identity, both responses mistakenly attribute the book to authors other than the real one, J.K. Rowling. Instead, they incorrectly name Chris Moriarty and Mark P. Swift as the authors.

% For the first block, the two questions both focus on summarizing the contents of the book ``Harry Potter and the Chamber of Secrets''. Even though we didn't ask about the author, both answers provide an author name. Neither of the names is the true author, J.K. Rowling. Instead, two fake names, Chris Moriarty and Mark P. Swift, are provided, respectively.

In the second block, we prompt the model to detail the relationships between three central characters from the book: Lily Potter, James Potter, and Sirius Black. The first response portrays James Potter and Sirius Black as a pair of detectives sharing a close bond. The second response introduces Lily Potter as a businessman's daughter and the love interest of the main character, Harry R. Potter. In the third response, Lily Potter and Sirius Black are depicted as former agents who share a professional connection. Interestingly, the model fabricates three different book series in its answers: the Harry Snow series, the Harry R. Potter series, and the Harry Ransom series, showcasing its creative yet inaccurate storytelling.

% In the second block, we ask the model to provide the relationships among three key characters in the book, namely, Lily Potter, James Potter, and Sirius Black. In the first answer, James Potter and Sirius Black are described as two detectives with a close relationship. In the second answer, Lily Potter is the daughter of a businessman and the love interest of the protagonist, Harry R. Potter. In the third answer, Lily Potter and Sirius Black are two former agents with a close professional relationship. What's worth it is that three book series are made up, which are the Harry Snow series, the Harry R. Potter series, and the Harry Ransom series. 

In the third block, we encounter two questions concerning the authorship of the Harry Potter series. The first answer incorrectly names Michael Morpurgo as the author, instead of J.K. Rowling. More intriguingly, the second response goes as far as to deny the very existence of the Harry Potter series.

% The third block presents the two questions asking about the author of the Harry Potter book series. The first author provided a fake name, Michael Morpurgo, to replace J.K. Rowling. Interestingly, the second answer directly denied the existence of the Harry Potter book series. 

Collectively, Table.~\ref{tab:contraditory} illustrate how the model generates inconsistent content across different dimensions: factual details about the book, the relationships between characters, and information about the author. Such inconsistencies signal that the model has not been effectively unlearned or fine-tuned for the task at hand.

% All three blocks provide examples of generating inconsistent content from the perspectives of book review (facts of the book), character relationship (book contents), and authors (book auxiliary information). Inconsistency in generated content is an indication that the model is not well-unlearned or fine-tuned for this task. 

\section{Insights into Unlearning of LLM}
\label{sec:insights}

In this section, we gently delve into several insights on the process of unlearning in LLMs, with the sincere hope that this information could be of assistance and inspiration for our readers' contemplation and future research endeavors.

\begin{itemize}
    \item \textbf{Ensuring Exhaustive Unlearning}: One of the primary challenges is ensuring that the unlearning process is thorough and leaves no residual knowledge of the target data. This task is particularly difficult due to the complexity and extensive depth of knowledge that is embedded in LLMs.
    
    \item \textbf{Balancing Unlearning and Retention}: Another critical challenge is maintaining a balance where the model unlearns specific information without significantly impacting its overall performance and capabilities. %It's vital that the unlearning process does not affect the model's ability to perform tasks that are unrelated. 
    For instance, when using specifically designed prompts to test the unlearning model for ``who-is-harry-potter,'' many prompts are met with side-effect outputs, such as meaningless responses, fake words, or even line break characters. This is also known as the disinformation of LLM. 
    
    \item \textbf{Hallucination after Unlearning}: It is noticeable that unlearned LLMs may have issues of under-unlearning or inconsistent generated contents. These problems are rooted in the hallucination of LLMs with a high chance. 
    
    \item \textbf{Adaptability to Different Types of Data}: Unlearning methods that are successful with certain types of data, like fictional narratives, may not be as effective with other types, such as factual or scientific data. A more complex challenge lies in developing unlearning methods that are versatile and effective across a wide range of data types. For instance, even if we are required to unlearn Newton's Laws of Motion, we still cannot change people's relevant experiences, e.g., a commercial plane accelerates along the runway and takes off.
    
    \item \textbf{Verification of Unlearning Unstructured Data}: Creating comprehensive and effective methods for evaluating the success of unlearning poses a further challenge. This involves developing benchmarks and tests to assess the degree of unlearning. Furthermore, confirming the unlearning of unstructured data, such as texts, presents even greater challenges compared to unlearning structured data. As this preliminary research shows, several aspects, including model integrity, over-unlearning, under-unlearning, and consistency of generated content, should be thoroughly considered in the verification process. 
\end{itemize}

\section{Summary}
\label{sec:summary}

The thorough analysis of unlearning in LLMs highlights the complexity of the issue. 
Unlearning structured data aims to enhance model accuracy and reduce bias, while unlearning unstructured data seeks to remove specific, often problematic, data from the model's knowledge base. The papers reviewed present novel approaches and encouraging results in both areas. However, challenges persist, especially in preserving the functional aspects of the model while achieving thorough unlearning. Future research should focus on enhancing these techniques to be more efficient and adaptable for a broad range of unlearning needs in LLMs. The ultimate goal is to develop LLMs that are not only powerful in language understanding and generation but also responsible, ethical, and in line with evolving legal standards.

%This comprehensive examination of unlearning in LLMs underscores the multifaceted nature of the challenge. While Classification Unlearning focuses on model accuracy and bias reduction, Knowledge Unlearning aims at removing specific, often problematic, data from the model's knowledge base. The reviewed papers demonstrate innovative methods and promising results in both categories. However, challenges remain, particularly in ensuring comprehensive unlearning without compromising the model's overall functionality. Future research should aim to refine these techniques, making them more efficient, adaptable, and capable of addressing a wider range of unlearning requirements in LLMs. The ultimate goal is to develop LLMs that are not only powerful in their language understanding and generation capabilities but also responsible, ethical, and compliant with evolving legal standards.

\iffalse
\section*{Acknowledgment}
\fi

\iffalse

\appendices
\section{Proof of the First Zonklar Equation}
Appendix one text goes here.

\section{}
Appendix two text goes here.

\fi


\ifCLASSOPTIONcaptionsoff
  \newpage
\fi

\bibliographystyle{IEEEtran}
\bibliography{nm}

\begin{thebibliography}{10}
\providecommand{\url}[1]{#1}
\csname url@samestyle\endcsname
\providecommand{\newblock}{\relax}
\providecommand{\bibinfo}[2]{#2}
\providecommand{\BIBentrySTDinterwordspacing}{\spaceskip=0pt\relax}
\providecommand{\BIBentryALTinterwordstretchfactor}{4}
\providecommand{\BIBentryALTinterwordspacing}{\spaceskip=\fontdimen2\font plus
\BIBentryALTinterwordstretchfactor\fontdimen3\font minus
  \fontdimen4\font\relax}
\providecommand{\BIBforeignlanguage}[2]{{%
\expandafter\ifx\csname l@#1\endcsname\relax
\typeout{** WARNING: IEEEtran.bst: No hyphenation pattern has been}%
\typeout{** loaded for the language `#1'. Using the pattern for}%
\typeout{** the default language instead.}%
\else
\language=\csname l@#1\endcsname
\fi
#2}}
\providecommand{\BIBdecl}{\relax}
\BIBdecl

\bibitem{singhal2023large}
K.~Singhal, S.~Azizi, T.~Tu, S.~S. Mahdavi, J.~Wei, H.~W. Chung, N.~Scales,
  A.~Tanwani, H.~Cole-Lewis, S.~Pfohl \emph{et~al.}, ``Large language models
  encode clinical knowledge,'' \emph{Nature}, vol. 620, no. 7972, pp. 172--180,
  2023.

\bibitem{cao2015towards}
Y.~Cao and J.~Yang, ``Towards making systems forget with machine unlearning,''
  in \emph{2015 IEEE symposium on security and privacy}.\hskip 1em plus 0.5em
  minus 0.4em\relax IEEE, 2015, pp. 463--480.

\bibitem{zhang2023right}
D.~Zhang, P.~Finckenberg-Broman, T.~Hoang, S.~Pan, Z.~Xing, M.~Staples, and
  X.~Xu, ``Right to be forgotten in the era of large language models:
  Implications, challenges, and solutions,'' \emph{arXiv preprint
  arXiv:2307.03941}, 2023.

\bibitem{jang2022knowledge}
J.~Jang, D.~Yoon, S.~Yang, S.~Cha, M.~Lee, L.~Logeswaran, and M.~Seo,
  ``Knowledge unlearning for mitigating privacy risks in language models,''
  \emph{arXiv preprint arXiv:2210.01504}, 2022.

\bibitem{eldan2023s}
R.~Eldan and M.~Russinovich, ``Who's harry potter? approximate unlearning in
  llms,'' \emph{arXiv preprint arXiv:2310.02238}, 2023.

\bibitem{karamolegkou2023copyright}
A.~Karamolegkou, J.~Li, L.~Zhou, and A.~S{\o}gaard, ``Copyright violations and
  large language models,'' in \emph{Proceedings of the 2023 Conference on
  Empirical Methods in Natural Language Processing}, 2023, pp. 7403--7412.

\bibitem{pawelczyk2023context}
M.~Pawelczyk, S.~Neel, and H.~Lakkaraju, ``In-context unlearning: Language
  models as few shot unlearners,'' \emph{arXiv preprint arXiv:2310.07579},
  2023.

\bibitem{min2023recent}
B.~Min, H.~Ross, E.~Sulem, A.~P.~B. Veyseh, T.~H. Nguyen, O.~Sainz, E.~Agirre,
  I.~Heintz, and D.~Roth, ``Recent advances in natural language processing via
  large pre-trained language models: A survey,'' \emph{ACM Computing Surveys},
  vol.~56, no.~2, pp. 1--40, 2023.

\bibitem{lu2022quark}
X.~Lu, S.~Welleck, J.~Hessel, L.~Jiang, L.~Qin, P.~West, P.~Ammanabrolu, and
  Y.~Choi, ``Quark: Controllable text generation with reinforced unlearning,''
  \emph{Advances in neural information processing systems}, vol.~35, pp.
  27\,591--27\,609, 2022.

\bibitem{chen2023unlearn}
J.~Chen and D.~Yang, ``Unlearn what you want to forget: Efficient unlearning
  for llms,'' \emph{arXiv preprint arXiv:2310.20150}, 2023.

\bibitem{wang2023kga}
L.~Wang, T.~Chen, W.~Yuan, X.~Zeng, K.-F. Wong, and H.~Yin, ``Kga: A general
  machine unlearning framework based on knowledge gap alignment,'' \emph{arXiv
  preprint arXiv:2305.06535}, 2023.

\bibitem{yu2023unlearning}
C.~Yu, S.~Jeoung, A.~Kasi, P.~Yu, and H.~Ji, ``Unlearning bias in language
  models by partitioning gradients,'' in \emph{Findings of the Association for
  Computational Linguistics: ACL 2023}, 2023, pp. 6032--6048.

\bibitem{kassem2023preserving}
A.~Kassem, O.~Mahmoud, and S.~Saad, ``Preserving privacy through
  dememorization: An unlearning technique for mitigating memorization risks in
  language models,'' in \emph{Proceedings of the 2023 Conference on Empirical
  Methods in Natural Language Processing}, 2023, pp. 4360--4379.

\bibitem{yao2023large}
Y.~Yao, X.~Xu, and Y.~Liu, ``Large language model unlearning,'' \emph{arXiv
  preprint arXiv:2310.10683}, 2023.

\bibitem{ma2022learn}
Z.~Ma, Y.~Liu, X.~Liu, J.~Liu, J.~Ma, and K.~Ren, ``Learn to forget: Machine
  unlearning via neuron masking,'' \emph{IEEE Transactions on Dependable and
  Secure Computing}, 2022.

\end{thebibliography}

\end{document}